\documentclass[sigconf,screen]{acmart}
\usepackage[ruled,vlined,linesnumbered]{algorithm2e}
\usepackage{makecell}
\usepackage{pifont}
\usepackage{amsmath}

\newcommand{\cmark}{\ding{51}}
\newcommand{\xmark}{\ding{55}}
\usepackage{wrapfig}
\usepackage{enumitem}

\AtBeginDocument{%
  }

\setcopyright{cc}
\setcctype{by}
\acmDOI{10.1145/3832783.3834399}
\acmYear{2026}
\copyrightyear{2026}
\acmISBN{979-8-4007-2882-2/2026/10}
\acmConference[ASE '26]{Proceedings of the 41st IEEE/ACM International Conference on Automated Software Engineering}{October 12--16, 2026}{Munich, Germany}
\acmBooktitle{Proceedings of the 41st IEEE/ACM International Conference on Automated Software Engineering (ASE '26), October 12--16, 2026, Munich, Germany}
\acmSubmissionID{ase26main-p1668-p}
\received{2026-03-26}
\received[accepted]{2026-06-18}

\begin{document}

\title{Towards Fully Automated Medical Imaging Code Generation via Validation-Based Context Engineering}

\author{Zixiao Zhao}
\orcid{0009-0002-0067-3101}
\affiliation{%
  \institution{University of Auckland}
  \city{Auckland}
  \country{New Zealand}
}
\email{zixiao.zhao@auckland.ac.nz}

\author{Jing Sun}
\orcid{0000-0002-1979-6622}
\affiliation{%
  \institution{University of Auckland}
  \city{Auckland}
  \country{New Zealand}
}
\email{jing.sun@auckland.ac.nz}

\author{Zhe Hou}
\orcid{0000-0001-7164-0580}
\affiliation{%
  \institution{Griffith University}
  \city{Brisbane}
  \country{Australia}
}
\email{z.hou@griffith.edu.au}

\author{Cheng-Hao Cai}
\orcid{0000-0001-6815-9091}
\affiliation{%
  \institution{Suzhou Industrial Park Monash Research Institute of Science and Technology}
  \city{Suzhou}
  \country{China}
}
\email{chenghao.cai.cs@gmail.com}

\author{Qian Liu}
\orcid{0000-0002-3162-935X}
\affiliation{%
  \institution{University of Auckland}
  \city{Auckland}
  \country{New Zealand}
}
\email{liu.qian@auckland.ac.nz}

\author{Mengze Li}
\orcid{0009-0000-2839-3844}
\affiliation{%
  \institution{University of Auckland}
  \city{Auckland}
  \country{New Zealand}
}
\email{mli840@aucklanduni.ac.nz}

\author{Zijian Zhang}
\orcid{0000-0002-6313-4407}
\affiliation{%
  \institution{Beijing Institute of Technology}
  \city{Beijing}
  \country{China}
}
\email{zhangzijian@bit.edu.cn}
\authornote{Corresponding author}

\author{Jin Song Dong}
\orcid{0000-0002-6512-8326}
\affiliation{%
  \institution{National University of Singapore}
  \city{Singapore}
  \country{Singapore}
}
\email{dcsdjs@nus.edu.sg}

\renewcommand{\shortauthors}{Zhao et al.}

\begin{abstract}
Large language models (LLMs) have demonstrated considerable promise in program generation for small-scale and conventional application development; however, they remain limited when applied to complex, domain-specific tasks such as medical image processing. General-purpose models lack explicit domain knowledge and robust validation mechanisms to ensure correctness, often requiring substantial human intervention to produce reliable processing pipelines. To address these limitations, we propose AutoMedImg, a multi-agent framework for fully automated medical image processing code generation. AutoMedImg orchestrates specialised agents across two phases: a Planning Phase that performs dataset analysis and architecture design with semantic and formal verification, and a Coding Phase that generates modules in parallel with static checking, execution testing, and assembly validation. This multi-stage validation mitigates error propagation throughout generation, while comprehensive auto-context engineering combining domain-specific knowledge bases, shared memory, and validation feedback automates context construction without manual prompting. A cross-project adaptive pipeline synthesis mechanism further accumulates validated pipelines and retrieves proven components for new tasks based on project similarity, enhancing generation efficiency through cross-project learning. Extensive evaluation across six diverse and well-established medical imaging datasets with five backbone LLMs demonstrates that AutoMedImg achieves zero human intervention , with Dice scores of up to 0.90 for segmentation tasks and 99\% accuracy for classification.
\end{abstract}

\ccsdesc[500]{Software and its engineering~Automatic programming}
\ccsdesc[500]{Computing methodologies~Multi-agent systems}
\ccsdesc[300]{Applied computing~Health care information systems}

\keywords{Automatic Programming, Large Language Models, Medical Imaging}


\maketitle

\section{Introduction}
The emergence of large language models (LLMs) has revolutionised the field of software engineering~\cite{jin2024llms,jiang2024survey,he2025llm}. 
Trained on massive code datasets, LLMs show impressive capabilities in natural language requirements comprehension~\cite{arora2024advancing}, relevant codebases search~\cite{li2024rewriting}, reasoning on software structures~\cite{ma2023lms}, and demonstrating substantial code generation capabilities~\cite{coignion2024performance}. 
On function-level code generation benchmarks such as HumanEval~\cite{chen2021evaluating} and MBPP~\cite{austin2021program}, state-of-the-art LLMs have achieved accuracy rates of 97.6\% and 92.7\%, respectively~\cite{HumanEvalLeaderboard,MBPPLeaderboard}. 
On SWE-bench~\cite{jimenez2024swebench}, a benchmark on solving real-world Github repository issues, LLMs have achieved a 76.8\% resolution rate~\cite{SWEbenchLeaderboard}. 
These results highlight the significant potential of LLMs in software engineering.

However, these benchmarks only demonstrate the capability of LLM systems in small-scale code generation or code repair within existing repositories. 
When generating complex code from scratch, current LLMs face significant limitations: (1) For complex problems, users must provide highly detailed prompts encompassing all task-relevant background information to achieve satisfactory results~\cite{li2024acecoder}; 
(2) Due to the hallucination problem inherent in LLMs and the lack of effective validation mechanisms, human intervention remains necessary during code generation~\cite{shin2025prompt}; 
(3) Without built-in validation mechanisms, LLMs cannot maintain consistent context and memory across code generation projects, even when successfully generating complete code~\cite{tao2404survey}. 
Medical image processing code generation significantly exemplifies these challenges. 
Medical images exhibit complex properties with multimodal data (CT, MRI), 2D/3D volumetric structures, various intensity ranges, and specialised formats (NIfTI, DICOM)~\cite{li2016first,mildenberger2002introduction}, demanding domain expertise that significantly exceeds what can be conveyed through manual prompting. 
Domain reliability requirements mean that hallucinated preprocessing steps or incorrect architectural choices are unacceptable, necessitating rigorous validation throughout generation. 
Complete pipelines span multiple interdependent stages from data loading through inference, while recurring patterns across projects demand cross-project knowledge retention that standard LLMs cannot exploit.
Automating this process would enable practitioners lacking deep learning expertise to deploy validated medical imaging pipelines, broadening access to AI-assisted diagnostics.
Moreover, established benchmarking datasets with ground truth annotations and standardised metrics enable objective assessment of generated pipelines. 
Together, these properties make medical image processing both a practically impactful and methodologically rigorous domain for evaluating automated code generation in complex, real-world settings.

Although several approaches attempt to tackle these challenges, they remain insufficient. 
Existing code generation frameworks or tools~\cite{hong2023metagpt,zhao2024mactg,cursor,github} lack domain-specific knowledge integration and systematic validation mechanisms, making human intervention indispensable for specialised domains.
Recent AutoML and multi-agent approaches~\cite{trirat2024automl,feng2025m} automate broader ML pipelines but remain confined or pre-defined datasets, requiring manual effort to extend to specialised domains or datasets.
Medical imaging automation methods~\cite{wang2024mednas,isensee2021nnu,MedSAM} incorporate domain expertise but address only partial pipeline components without achieving fully automated generation. 
No existing approach combines comprehensive domain knowledge, systematic validation, and consistent context management to enable fully automated generation of complete, validated pipelines from scratch.

To address these limitations, we propose AutoMedImg, a multi-agent framework for automated medical image processing code generation via validation-based context engineering. 
AutoMedImg integrates domain knowledge bases eliminating exhaustive manual prompting, systematic multi-stage validation preventing error propagation, and adaptive pipeline synthesis maintaining context across projects. 
Built on a multi-agent architecture, AutoMedImg orchestrates the pipeline from dataset analysis through architecture design, parallel module implementation, and final assembly. 
Evaluation on six medical imaging datasets demonstrates that AutoMedImg achieves fully automated code generation while maintaining competitive model performance.

The main contributions of this work are as follows:
\begin{enumerate}
    \item We propose a \textbf{validation-based context engineering approach} integrating medical imaging domain knowledge, shared memory, and multi-stage validation to automatically construct and maintain generation context, enabling error detection and correction without manual prompting.
    \item We develop an \textbf{adaptive pipeline synthesis mechanism} that transfers validated medical imaging pipeline components across projects, enabling efficient knowledge reuse while maintaining pipeline quality.
    \item We present AutoMedImg, a multi-agent framework for \textbf{fully automated medical image processing code generation}, evaluated on six diverse datasets, achieving zero human intervention with competitive model performance.
\end{enumerate}

The remainder of this paper is organised as follows: 
Section~\ref{sec:related} reviews related work on automated code generation and medical imaging automation. 
Section~\ref{sec:method} presents the AutoMedImg framework, detailing the multi-agent architecture, systematic validation mechanisms, and validation-guided context engineering.
Section~\ref{sec:experiments} evaluates existing state-of-the-art code generation approaches on medical image processing to identify their limitations, and demonstrates how AutoMedImg's validation-based design addresses these limitations across six medical imaging datasets with ablation studies. 
Section~\ref{sec:discussion} presents a qualitative comparison with medical imaging automation methods and discusses threats to validity. 
Section~\ref{sec:conclusion} concludes with a summary and future research directions.

\section{Related Work}
\label{sec:related}

\paragraph{Automated code generation with LLMs.}
LLM-based code generation has evolved from early models like InCoder~\cite{fried2022incoder} and CodeGen~\cite{nijkamp2022codegen} to proprietary models such as GPT~\cite{openai_introducing_2025}, Claude~\cite{claude4.5opus}, and Gemini~\cite{gemini-3-pro} achieving state-of-the-art performance on benchmarks like HumanEval~\cite{chen2021evaluating} and MBPP~\cite{austin2021program}. Self-debugging frameworks~\cite{chen2023teaching} and execution-feedback methods~\cite{ni2024next} enable refinement based on runtime errors but rely on human-provided test cases. Multi-agent frameworks such as ChatDev~\cite{qian2024chatdev} and AgentCoder~\cite{huang2023agentcoder} introduced dynamic test generation through agent collaboration, yet remain limited to function-level or module-level generation. Contemporary tools including GitHub Copilot~\cite{github}, Cursor~\cite{cursor}, and Antigravity~\cite{antigravity} integrate LLMs into IDEs with execution capabilities but still require substantial human intervention across complex multi-stage tasks in specialised domains.

\begin{figure*}[!htbp]
    \centering
    \includegraphics[width=0.9\linewidth]{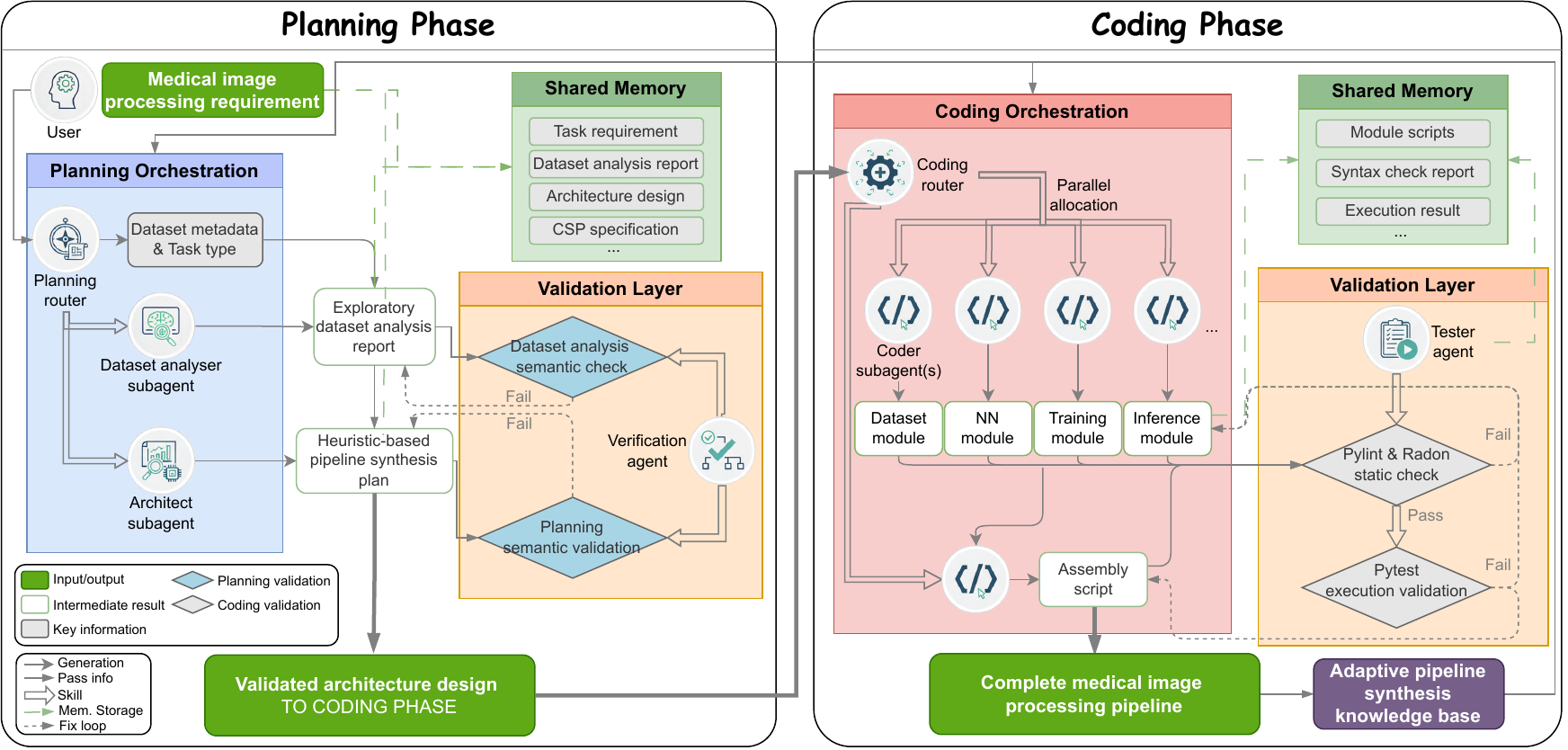}
    \vspace{-8pt}
    \caption{AutoMedImg's two-phase workflow: Planning and Coding }
    \label{fig:workflow}
    \Description{vace workflow}
\end{figure*}

\paragraph{Automation in medical image processing.}
In the field of medical imaging, numerous automated algorithm design and optimization methods have been developed. 
NAS~\cite{liu2021survey,ren2021comprehensive} automatically discovers optimal network structures but requires predefined search spaces and manual pipeline implementation~\cite{wang2024mednas}. nnU-Net~\cite{isensee2021nnu} automates segmentation pipeline configuration but requires manual environment setup and is restricted to segmentation tasks. Foundation models like SAM~\cite{kirillov2023segment,ravi2024sam} and MedSAM~\cite{MedSAM} eliminate task-specific training but require interactive prompts during inference and address only segmentation. 
AutoML-Agent~\cite{trirat2024automl} automates full-pipeline ML across diverse tasks but requires manually provided skeleton code for new task types and relies solely on prompt-based rather than execution-grounded verification.
DS-Agent~\cite{guo2024ds} leverages case-based reasoning for data science automation but lacks medical imaging knowledge for generating appropriate pipelines, and its sole reliance on execution feedback can only guarantee runnable code without ensuring alignment with domain conventions.
VoxelPrompt~\cite{hoopes2024voxelprompt} integrates a jointly-trained vision-language agent for end-to-end radiological analysis but focuses exclusively on inference-time task execution without addressing automated pipeline development or code generation. 
M$^3$Builder~\cite{feng2025m} automates medical imaging ML workflows through multi-agent coordination but restricts generation to pre-defined datasets with accompanying code templates, requiring manual preparation for new datasets. 
These approaches address only partial pipeline components, requiring substantial expertise for deployment, and none achieves fully automated generation of complete pipelines from scratch.

\section{Method: AutoMedImg Framework}
\label{sec:method}
In this section, we present AutoMedImg workflow and framework through 3 components: 
Section~\ref{sec:agent} details agent specifications and context integration mechanisms; 
Section~\ref{sec:context} explains validation-based context engineering in the Planning and Coding phases; 
Section~\ref{sec:adaptive} presents cross-project adaptive pipeline synthesis.

Medical imaging pipelines require understanding the dataset characteristics before implementation. 
AutoMedImg addresses this through a two-phase architecture, Planning and Coding, as illustrated in Figure~\ref{fig:workflow}.
\textbf{Planning Phase.} Given a medical image processing requirement, the Planning Router first invokes the Dataset Analyser subagent to perform exploratory dataset analysis. 
The Verification Agent validates this analysis through semantic checking, ensuring all required dataset features are extracted; failures trigger re-analysis with diagnostic feedback. 
Next, the Architect subagent synthesises a pipeline design based on validated dataset characteristics. 
This design undergoes planning semantic validation at two levels: contextual validation checks architecture-dataset compatibility, modularity requirements, and medical imaging conventions; formal validation converts the architecture to CSP\# specification for PAT-based model checking, ensuring process correctness.
Throughout this process, intermediate results (dataset analysis report, architecture design, CSP specification) are stored in Shared Memory for agent coordination. 
Only architectures passing semantic validation propagate to the Coding Phase as validated pipeline designs.
\textbf{Coding Phase.} The Coding Router decomposes the validated pipeline design and allocates multiple Coder subagents for parallel module generation (Dataset, NN, Training, Inference modules). 
The Tester Agent validates each module through static checking (Pylint, Radon) and execution testing (Pytest); failures trigger code refinement with diagnostic feedback. 
Validated modules are integrated into an assembly script, which is then subjected to final end-to-end validation. 
Successfully validated pipelines that achieve satisfactory performance are entered into the Adaptive Pipeline Synthesis Knowledge Base, along with their project characteristics and performance metrics, enabling future projects to retrieve and adapt proven components from similar tasks.
This two-phase architecture with systematic multi-stage validation enables fully automated generation of production-ready medical imaging pipelines.
\subsection{AutoMedImg agent specification}
\label{sec:agent}
\subsubsection{Agent Coordination Structure}
Complex medical image processing tasks require comprehensive expertise over multiple domains, including medical image data analysis, library dependencies and APIs, data loading strategies, and neural network architecture design. 
A single-agent approach faces fundamental limitations: (1) simultaneously invoking knowledge across diverse domains creates confusion and inconsistent reasoning; (2) maintaining accurate context becomes increasingly difficult as context accumulates through planning, validation, and coding phases; (3) single-agent systems cannot exploit parallel execution for efficiency improvement. 

AutoMedImg addresses these limitations through a multi-agent architecture organised as a router-subagent hierarchy with an independent Validation Layer.
To distribute domain expertise, AutoMedImg employs specialised agents: the Planning Router invokes Dataset Analyser and Architect, while the Coding Router allocates multiple Coder subagents in parallel for concurrent module generation, each possessing focused knowledge rather than handling multiple domains simultaneously.
Each agent maintains its own context window, with Shared Memory storing validated intermediate results that agents retrieve as needed, preventing context accumulation in a single window.
Critically, the Verification Agent and Tester Agent operate within an independent Validation Layer rather than embedded within generating agents, eliminating self-validation bias.
Routers coordinate workflow progression based on validation outcomes: only verified results advance to subsequent stages, while failures trigger automated correction loops with diagnostic feedback. 
This architecture addresses domain knowledge specialisation, phase-specific context management, and parallel execution requirements.

\begin{figure}[!htbp]
\centering
    \includegraphics[width=0.9\linewidth]{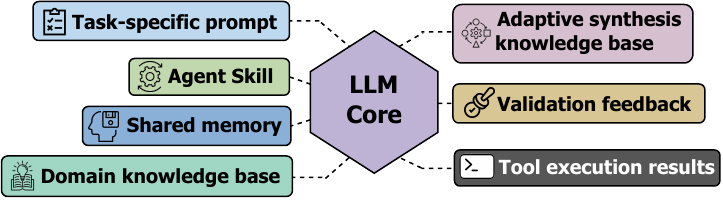}
    \vspace{-8pt}
    \caption{AutoMedImg Agent Context Integration}
    \label{fig:agent_arch}
    \Description{AutoMedImg Agent Tool and Context Integration}
\end{figure}

\subsubsection{Agent Context Integration}
AutoMedImg achieves comprehensive agent context engineering through seven core components, as illustrated in Figure~\ref{fig:agent_arch}.
Task-specific prompts define each agent's objectives and expected outputs within the workflow. 
Agent Skills provide structured execution capabilities that guide agents in performing specific tasks step-by-step with precise instructions.
Domain knowledge bases supply agent-specific expertise for generation and validation.
Shared memory stores validated intermediate results accessible across agents.
Adaptive synthesis knowledge base maintains validated pipelines from previous projects.
Validation feedback provides diagnostic information guiding iterative corrections.
Tool execution results capture outputs from command-line or Model Context Protocol (MCP) tools invoked during execution.

Specific resource assignments vary by agent role. 
Planning Router and Coding Router employ orchestration Skills for workflow coordination: Planning Router sequentially invokes Dataset Analyser and Architect subagents, while Coding Router concurrently allocates multiple Coder subagents for parallel module generation. 
Dataset Analyser, Architect, and Coder each access specialised knowledge bases: The Dataset Analyser's knowledge base is derived from nnU-Net~\cite{isensee2021nnu} dataset card specifications, capturing standardised medical imaging dataset properties; The Architect's knowledge base is manually curated from leading neural network architectures on \href{https://grand-challenge.org/}{Grand Challenge}, providing validated pipeline design patterns across diverse medical imaging tasks, with CSP\# modeling Skill for converting architecture designs to formal specifications; The Coder's knowledge base is extracted from official MONAI API documentation, supplying library-specific implementation templates.
The Verification Agent's knowledge base comprises manually summarised ISO/IEC 25023 modularity standards and IEC 62304 medical device software conventions, providing structured validation criteria and scoring rubrics for semantic assessment, CSP\# interpretation knowledge base for understanding verification outputs, alongside Skills for invoking PAT tool for formal verification.
Tester accesses the Testing Knowledge Base containing test case patterns and coverage criteria, while utilising static analysis tools (Pylint, Radon) and dynamic testing frameworks (Pytest) through MCP. 
All agents share access to shared memory for validated intermediate results and adaptive synthesis knowledge base for cross-project reference.
Validation feedback from failed checks and tool execution results is incorporated during the fix-loop, enabling targeted corrections.
This comprehensive context integration enables each agent to automatically construct domain-appropriate context without manual prompting.

\subsection{Validation-based context engineering}
\label{sec:context}
Traditional LLM-based code generation operates through prompt-and-generate cycles, where agents produce outputs solely based on input prompts and pre-trained knowledge. 
This approach suffers from three critical limitations: (1) without checking outputs, agents cannot assure output quality and reliability; (2) outputs propagate unchecked, allowing errors to cascade through subsequent stages, and (3) without memory management through systematic validation, high-quality patterns may not be reliably adapted or reused across projects. 
In medical image processing, these limitations manifest as misaligned preprocessing sequences, inconsistent architecture designs, and functionally incorrect implementations.

AutoMedImg addresses these limitations through validation-based context propagation: only outputs that satisfy validation criteria propagate to subsequent agents, while failures generate diagnostic feedback, enabling targeted refinement. 
The mechanism operates across two phases: the Planning Phase validates dataset analyses and architecture designs through dataset analysis and planning semantic validation; the Coding Phase validates module implementations and system assembly through static and execution validation. 
We formalise this mechanism through the following definitions:

\textbf{Definition 1 (Agent Context).} The context of agent $A_i$ is defined as:
\begin{equation}
C_{A_i} = \langle P_i, M, K_D^i, K_{Ada}, H_i, V_i \rangle
\label{eq:context}
\end{equation}
where $P_i$ represents agent $A_i$'s task-specific prompt; $M$ denotes shared memory containing accumulated validated outputs from all predecessor agents; $K_D^i$ is agent $A_i$'s domain knowledge base; $K_{Ada}$ is the Adaptive Pipeline Synthesis Knowledge Base; $H_i$ is agent $A_i$'s tool or Skill execution history; and $V_i$ is the agent's accumulated validation feedback.

\textbf{Definition 2 (Context Evolution).} Agent $A_i$ generates output $o_i = g_{A_i}(C_{A_i})$ from its context, which is then evaluated by validation function $\mathcal{V}: \mathcal{O} \rightarrow \{\texttt{pass}, (\texttt{fail}, f)\}$. On pass, $o_i$ is stored in shared memory $M$ and the next agent $A_{i+1}$ proceeds with updated context; on fail, diagnostic feedback $f_i$ is appended to $V_i$ and $A_i$ retries with enriched context, enabling targeted correction.

\subsubsection{Domain Knowledge and Shared Memory}
Domain knowledge bases $K_D^i$ and shared memory $M$ serve as foundational context components enabling validation-based context engineering across both Planning and Coding phases. 
Domain knowledge provides agent-specific expertise without requiring manual specification, while shared memory enables the propagation of validated information without redundant computation.

\textbf{Domain Knowledge for Expertise Provision.} Each agent accesses a domain knowledge base $K_D^i$ tailored to its role, providing specialised contextual knowledge during generation. 
Each agent's $K_D^i$ is constructed from general medical-domain standards and documentation as a knowledge graph using the Cognee framework~\cite{markovic2025optimizinginterfaceknowledgegraphs}, enabling natural language querying via MCP tool invocation during generation, and covering the most common medical imaging tasks (supervised/semi-supervised segmentation and classification, 2D/3D), modalities (CT, MRI, X-ray), and architectures.
For example, the Dataset Analyser's $K_D^i$ encodes properties such as: \textit{"CT images require HU windowing in range [-1000, 3000]; voxel spacing must be extracted and preserved for physical scale consistency."} 
The Architect's $K_D^i$ stores patterns such as: \textit{"3D anisotropic data requires physical spacing-aware resampling; multi-organ segmentation with class imbalance requires Dice + Cross-entropy combined loss."} 
The Verification Agent's $K_D^i$ contains criteria such as: \textit{"Ensure patient-level splitting to prevent data leakage; Validate CT preprocessing includes orientation normalisation before resampling."}
Without domain knowledge, the agent's outputs lack domain-specific constraints, leading to inappropriate outputs. 

\textbf{Shared Memory for Context Propagation.} Shared memory $M$ is implemented via Eion MCP tool, a persistent key-value store for multi-agent coordination, where validated outputs are written after each successful validation and retrieved through natural language queries, enabling flexible and precise access to verified information without redundant derivation.
For instance, after the Dataset Analyser completes validated analysis, $M$ contains: modality (CT), dimensions (512×512×196), voxel spacing (0.8×0.8×3.0mm), intensity range (-1000 to 3000 HU), and task type (supervised segmentation). The Architect retrieves these directly to design a patch-based 3D pipeline with HU windowing preprocessing. 
Similarly, validated module code propagates through $M$ to assembly, eliminating redundant validation and ensuring consistency across workflow stages.

\subsubsection{Planning Phase Context Engineering}
The Planning Phase constructs validated architecture designs through sequential agent invocation with validation-guided refinement. 
Dataset Analyser extracts medical imaging characteristics; Architect designs pipeline architectures grounded in validated dataset analysis; Verification Agent ensures planning correctness through two-level semantic validation: dataset analysis semantic checking verifies completeness and domain compliance; planning semantic validation checks architecture correctness through contextual domain conventions and formal CSP\# model checking.

\textbf{Dataset Analysis Semantic Check.} Semantic validation $\mathcal{V}_{s}$ verifies structural completeness and domain compliance of the dataset analysis:
\begin{equation}
\mathcal{V}_{s}(o_{DA}) = \texttt{pass} \Leftrightarrow
\forall p \in \mathcal{P}_{DA}: p \in o_{DA} \land \texttt{consistent}(o_{DA})
\end{equation}
where $\mathcal{P}_{DA} = \{modality, dimensions, spacing, intensity, count\}$ denotes the required dataset properties.
In practice, the Verification Agent queries its domain validation knowledge base and evaluates the dataset analysis JSON output, checking for the presence and consistency of each required property in $\mathcal{P}_{DA}$. 
For example, for a CT dataset, validation checks that modality is correctly identified, voxel spacing is extracted (e.g., 0.8×0.8×3.0mm), HU intensity range is reported (e.g., -1000 to 3000), and case counts are consistent. A missing spacing value or inconsistent case count triggers re-analysis with targeted diagnostic feedback.
This check is conducted by the Verification Agent through rule-based semantic assessment: the agent queries its domain validation knowledge, which contains structured validation criteria and scoring rubrics, systematically evaluates each architectural decision against these specifications, and generates detailed diagnostic reports that enable targeted refinement of failed components.

\textbf{Planning Semantic Validation.} 
Contextual semantic validation checks two criteria. 
$\texttt{compatible}(\cdot, M)$ verifies that the architecture aligns with validated dataset characteristics in shared memory — for instance, anisotropic volumetric data requires patch-based strategies and voxel-space resampling to preserve physical scale, while multi-organ segmentation requires loss functions that address class imbalance from organ volume distributions. 
On the other hand, $\texttt{valid\_design}(\cdot, K_D^{Arch})$ ensures compliance with ISO/IEC modularity requirements and medical imaging conventions drawn from the domain knowledge base, including HU windowing for CT, z-score normalisation for MRI, and patient-level data splitting to prevent leakage. 
For example, proposing a 2D UNet for a 3D CT dataset, or applying generic resize instead of voxel-space resampling, triggers a compatibility failure with diagnostic feedback specifying the violated convention, enabling targeted architecture revision.
This validation is executed in the same procedure as $\mathcal{V}_{s}$.

Formal semantic validation converts the architecture to a CSP\# specification via dedicated Skill and validates it via PAT tool:
\begin{equation}
\mathcal{V}_{fs}(o_{Arch}) = \texttt{PAT}(\texttt{CSP}(o_{Arch}))
\end{equation}
verifying deadlock-freedom, race condition absence, and LTL properties such as $\square(model\_initialized \rightarrow \lozenge training\_finished)$ and $\square(test\_data\_loaded \rightarrow \lozenge inference\_complete)$. Properties are derived from manually verified templates that specify standard safety and liveness requirements for medical imaging workflows. 
The Verification Agent adapts these templates to the current project through regex-based Skill invocation, substituting project-specific process names and parameters while preserving the verified safety and liveness properties.
Since medical image processing pipelines follow consistent algorithmic patterns (data loading, preprocessing, training, inference), the templates comprehensively cover the required properties across different tasks.

Through two-level semantic validation on architecture design, the Planning Phase achieves validation-based context engineering: each agent iteratively refines its outputs through validation feedback, and only verified outputs enter shared memory $M$ for subsequent agents, preventing error propagation while eliminating manual specification requirements.

\subsubsection{Coding Phase Context Engineering}
The Coding Phase translates the validated architecture design into executable code through validation-guided synthesis. 
Upon receiving $o_{Arch} \in M$ from the Planning Phase, the Coding Router parses the JSON-structured architecture design, which explicitly specifies class definitions with inheritance relationships, method signatures with type annotations, attributes with pre-defined values (e.g., learning rate, patch size), and inter-module dependencies. 
The Router then decomposes the design into module specifications for parallel Coder agents to implement.
This phase employs three validation mechanisms operating at increasing scopes: static checking validates the quality of individual module code; execution validation verifies the functional correctness of modules through dynamic testing; and assembly validation ensures the consistency of system-level integration.

\textbf{Static Checking on Module Code.} Static checking $\mathcal{V}_{stat}$ verifies syntactic correctness, dependency consistency, and code quality before execution. 
The Tester Agent employs Pylint for syntax validation and Radon for complexity measurement:
\begin{equation}
\mathcal{V}_{stat}(code_i) = \texttt{pass} \Leftrightarrow \bigwedge_{m \in \{Pylint, CC, MI\}} \texttt{threshold}(m, code_i, \tau_m)
\end{equation}
where $\texttt{Pylint}(\cdot)$ measures syntactic correctness and dependency validity, $CC(\cdot)$ denotes cyclomatic complexity, $MI(\cdot)$ represents maintainability index, and $\tau_m$ denotes the quality threshold for each metric $m$ gated by ISO/IEC 25023 standards. 
In practice, the Tester Agent invokes Pylint and Radon via MCP tool calls, collecting their structured outputs and evaluating CC and MI scores against predefined thresholds ($\tau_{CC} = \tau_{MI} = 80$), while Pylint pass/fail is determined automatically by its built-in rule set.
Static checking eliminates syntactic errors before execution, reducing the complexity of runtime validation. Failed checks enrich $V_{Coder}$ with diagnostic feedback including error locations, dependency conflicts, and API mismatches for targeted correction.

\textbf{Execution Validation on Module Code.} Execution validation $\mathcal{V}_{exec}$ verifies runtime correctness through dynamic testing via Pytest. The Tester generates test cases covering three categories: assumption tests validate input constraints (e.g., verifying input tensor shape matches expected dimensions $[B, C, D, H, W]$ for 3D volumetric data); functional tests verify computational correctness (e.g., checking that preprocessing output intensity range falls within the expected HU window after normalisation); and integration tests check interface compatibility (e.g., confirming that preprocessed output shape matches NN module input expectations). Formally:
\begin{equation}
\mathcal{V}_{exec}(code_i) = \bigwedge_{t \in Tests(code_i)} \texttt{execute}(t, code_i) = \texttt{expected}(t)
\end{equation}
The Tester Agent generates 30--40 test cases per module via LLM invocation, guided by testing rules and examples from its knowledge base. 
Test cases are executed via MCP tool calls to Pytest, with execution traces captured and appended to $V_{Coder}$ as diagnostic feedback for targeted correction.
Failed tests trigger refinement cycles where execution traces reveal runtime behaviour, enabling targeted debugging. Only modules satisfying both static and execution validation propagate to assembly.

\textbf{Assembly Validation.} Assembly validation $\mathcal{V}_{a}$ verifies system-level correctness after module integration by applying both static checking and execution validation to the assembled pipeline. While individual modules may pass isolated validation with mocked inputs, module-level correctness does not guarantee system-level correctness when real data flows among modules. For example, the Dataset module may correctly output a dictionary with key \texttt{image} that passes its own validation, while the NN module expects key \texttt{data}, violating the interface contract that only manifests when both modules are assembled. Assembly validation detects such interface mismatches, dimension incompatibilities, and incorrect sequencing, triggering targeted refinement with diagnostic feedback. 

Through static, execution, and assembly validation, the Coding Phase achieves multi-scale validation-based context engineering: individual modules refine through validation feedback, validated modules propagate through $M_{code}$, and only high-performing complete pipelines enter $K_{Ada}$ for future adaptive synthesis, ensuring both correctness and quality throughout implementation.

\subsection{Cross-project adaptive pipeline synthesis}
\label{sec:adaptive}
\begin{figure}[!htbp]
    \centering
    \includegraphics[width=\linewidth]{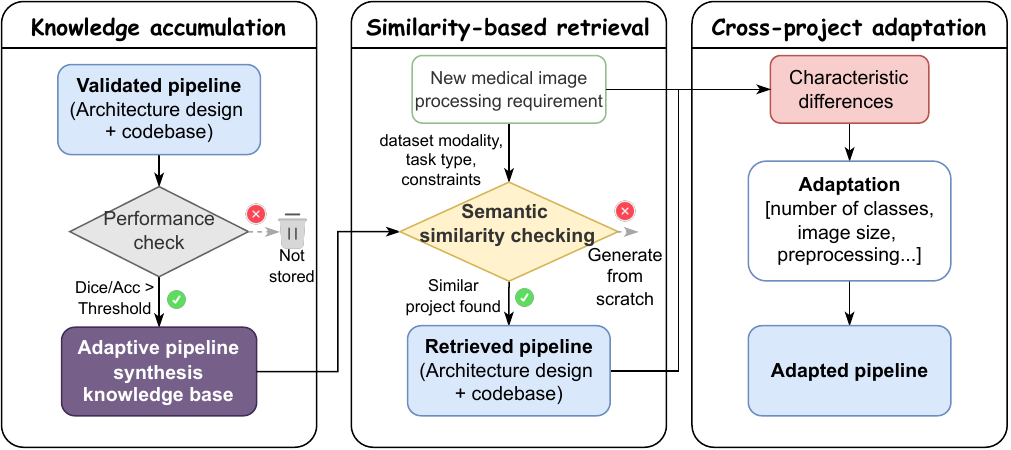}
    \caption{Cross-project Adaptive Pipeline Synthesis Workflow}
    \label{fig:ada_workflow}
    \Description{Cross-project Adaptive Pipeline Synthesis Workflow}
    \vspace{-8pt}
\end{figure}
A critical limitation of LLM-based code generation is the lack of cross-project continuity: each generation session operates independently, without retaining knowledge from previous successful tasks. 
In medical image processing, analogous scenarios frequently arise: analysing different organs within the same modality, segmenting tissues with similar features, or processing datasets with comparable structures. 
These recurring patterns necessitate cross-project learning that standard LLMs cannot exploit.
Cross-project adaptive pipeline synthesis addresses this by accumulating validated solutions in $K_{Ada}$ and leveraging them to guide future generations. 
This constrains the solution space from exhaustive assumption to synthesis within validated neighbourhoods, minimising error, and enhancing generation efficiency.
Figure~\ref{fig:ada_workflow} illustrates this three-stage process:

\textbf{Knowledge Accumulation.} Successfully validated pipelines achieving satisfactory performance enter $K_{Ada}$ for cross-project reference:
\begin{equation}
K_{Ada} \leftarrow K_{Ada} \cup \{(o_{Arch}, code, \xi, \rho)\} \quad \text{if } \mathcal{V}_{a} = \texttt{pass} \land \rho \geq \theta
\label{eq:kada_entry}
\end{equation}
where $o_{Arch}$ represents the validated architecture design, $code$ represents the complete pipeline codebase, $\xi$ captures project characteristics (dataset modality, task type, dimensionality, architectural constraints), $\rho$ denotes performance metrics (Dice score, accuracy), and $\theta$ defines the quality threshold. 
In practice, validated pipeline entries are appended incrementally to the agent knowledge graph via Cognee, enabling retrieval through the same natural language querying mechanism as the domain knowledge base; each dataset can only reference pipelines validated from previously processed datasets, ensuring all retrieved components originate exclusively from prior evaluation datasets and avoid information leakage across projects.
Each entry preserves the complete pipeline structure, validated components, and achieved performance, enabling future projects to access proven solutions rather than generating them from scratch.

\textbf{Similarity-based Retrieval.} When processing a new project with requirements $\xi_{new}$, AutoMedImg retrieves relevant pipelines from $K_{Ada}$ based on characteristic similarity:
\begin{equation}
\mathcal{R}(\xi_{new}) = \{(o_{Arch}, code, \xi, \rho) \in K_{Ada} \mid sim(\xi_{new}, \xi) \geq \tau_{sim}\}
\label{eq:retrieval}
\end{equation}
In practice, $sim(\cdot, \cdot)$ is computed via LLM-based semantic matching, where the Router evaluates characteristic alignment between the new project and retrieved entries through natural language reasoning over project descriptors, applying hierarchical matching: hard constraints, including dataset modality and task type, must match exactly, while soft constraints such as dimensionality and architectural requirements are assessed by degree of similarity. $\tau_{sim}$ defines the minimum alignment threshold across both hard and soft constraints, where any mismatch in hard constraints immediately disqualifies retrieval regardless of soft constraint similarity.
Retrieved pipelines serve as reference implementations, providing validated architecture designs and codebases for adaptation. 

\textbf{Cross-project Adaptation.} For a retrieved entry $(o_{Arch}, code, \\
\xi_{old}, \rho) \in \mathcal{R}(\xi_{new})$, agents extract components and adapt them based on characteristic differences $\Delta(\xi_{old}, \xi_{new})$ such as class numbers, image size, and preprocessing requirements, and adapt the retrieved pipeline accordingly. 
For example, when processing semi-supervised 3D CT segmentation after a supervised 3D CT segmentation task, AutoMedImg retrieves the segmentation pipeline as a reference, adapts the same data loading strategy, adjusts the number of output classes, and adds the unlabelled data handling logic, while retaining the validated preprocessing pipeline and 3D UNet architecture. 
All adapted outputs undergo validation in the main pipeline to ensure correctness for the target project.

Through $K_{Ada}$, AutoMedImg achieves cross-project adaptation: validated pipelines accumulate incrementally, relevant components are retrieved based on project similarity, and adaptations undergo validation ensuring correctness. This mechanism enables knowledge reuse while maintaining quality standards, operating without human supervision.

\section{Evaluation}
\label{sec:experiments}

To thoroughly investigate the effectiveness of AutoMedImg, we formulate the following research questions:

\textbf{RQ1: What are the fundamental limitations of existing code generation approaches when applied to medical image processing tasks?}  

\textbf{RQ2: How does AutoMedImg address the fundamental limitations of existing code generation approaches compared to advanced multi-agent and automated ML frameworks?}  

\textbf{RQ3: How do AutoMedImg's individual validation and context engineering components contribute to reducing LLM errors and ensuring code correctness?}  

\textbf{RQ4: How does AutoMedImg's multi-agent architecture improve code generation quality compared to applying the same context engineering techniques in a single-agent system?}  

This section proceeds as follows: Section~\ref{sec:datasets} presents the benchmark datasets and their features.
Section~\ref{sec:eval-setup} describes experimental configurations and evaluation metrics. 
Section~\ref{sec:results} reports experimental results organised by research question, including comprehensive ablation studies. 

\subsection{Benchmarking datasets}
\label{sec:datasets}
To evaluate the performance of AutoMedImg and baseline methods across different medical image processing sub-tasks, we selected four task categories comprising six datasets in total:\\
\textbf{FLARE22~\cite{ma2024unleashing}:} A \textit{semi-supervised 3D segmentation} dataset for 13 abdominal organs in CT scans (50 labelled, 2000 unlabelled cases, 512$\times$512$\times$196), challenging due to the large unlabelled pool with limited supervision.\\
\textbf{ACDC~\cite{bernard2018deep}:} A \textit{supervised 3D segmentation} dataset for three cardiac structures in MRI (215$\times$228$\times$10), challenging due to significant anisotropic voxel spacing.\\
\textbf{BTCV~\cite{landman2015miccai}:} A \textit{supervised 3D segmentation} dataset for 13 organs in CT scans (50 cases, 512$\times$512$\times$123), challenging due to limited samples and imbalanced organ volumes.\\
\textbf{FLARE21~\cite{ma2022fast}:} A \textit{supervised 3D segmentation} dataset for 4 abdominal organs in CT scans (361 cases, 512$\times$512$\times$126), representing a relatively simpler segmentation task than FLARE22.\\
\textbf{Promise12MSBench~\cite{kucs2024medsegbench}:} A \textit{supervised 2D segmentation} dataset for binary prostate segmentation in MRI (512$\times$512), derived from PROMISE12~\cite{litjens2014evaluation}.\\
\textbf{OrganMNIST3D~\cite{yang2023medmnist}:} A \textit{supervised 3D classification} dataset for 11 organ categories in CT scans (1,742 samples, 64$\times$64$\times$64), the most straightforward task in our benchmark.

These six datasets are selected because they cover the two most fundamental medical image processing tasks (segmentation and classification), are publicly available and widely adopted benchmarks, and span a comprehensive difficulty gradient from simple supervised classification to complex 3D semi-supervised segmentation, enabling robust evaluation across diverse scenarios.

\begin{figure*}[!htbp]
\centering
    \includegraphics[width=0.8\linewidth]{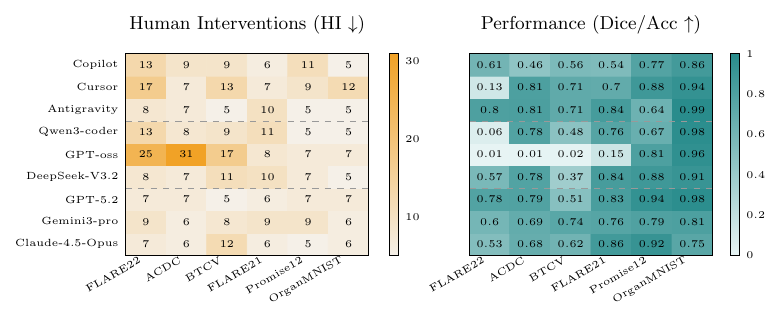}
    \vspace{-15pt}
    \caption{Performance of Code Generation Tools and Base LLMs on Medical Image Processing Tasks}
    \label{fig:baseline_evaluation}
    \Description{Performance of Code Generation Tools and Base LLMs on Medical Image Processing Tasks}
    \vspace{-10pt}
\end{figure*}

\subsection{Evaluation settings \& criteria}
\label{sec:eval-setup}
\subsubsection{Baseline Prompt Design and Configuration}
To evaluate the fundamental limitations of existing code generation approaches on medical image processing tasks, all baseline methods receive carefully designed prompts that provide equivalent task structure and necessary tool access, enabling them to attempt the same four-stage pipeline as AutoMedImg:
\textbf{Dataset Analysis:} Generate scripts to examine dataset file structure, image dimensions, modality characteristics and intensity distributions.
\textbf{Architecture Design:} Design the complete pipeline architecture based on dataset analysis results, including preprocessing and data-loading strategies, neural network structure, training configurations, and inference workflow.
\textbf{Module Implementation:} Implement individual modules according to the architecture design.
\textbf{Pipeline Assembly:} Generate an assembly script integrating all modules into an executable end-to-end pipeline.

Baselines also receive command-line execution permissions and file system access where their architecture permits, enabling them to: (1) execute Python scripts for runtime debugging, (2) read/write files during code generation, (3) check and install required dependencies, and (4) attempt to iteratively refine code based on execution feedback. This configuration allows us to identify what types of errors and limitations emerge without AutoMedImg's validation-based context engineering and multi-agent coordination.

\subsubsection{Experiment Environmental Settings}
The evaluation was performed on an Ubuntu 22.04 server equipped with an AMD Ryzen Threadripper PRO 3995WX CPU (64 cores), 252GB RAM, and 2 NVIDIA RTX A6000 GPUs (48GB VRAM each).
We provide a unified Python environment with pre-installed medical imaging libraries and common dependencies.
All methods can install additional dependencies as needed during code generation. 
Regarding AutoMedImg's domain knowledge, it's constructed from general medical-domain standards independent of the evaluation datasets selected for this study, ensuring no dataset-specific knowledge is embedded in the framework prior to evaluation.

\subsubsection{Evaluation Criteria}
We evaluate methods across three dimensions. To measure \textbf{autonomy}, we count human interventions (HI) --- the number of times a human must intervene to correct errors or provide guidance that the system cannot autonomously resolve, including error correction, design guidance, and workflow instructions. HI takes three forms: direct prompt interactions (providing instructions, relaying error messages, or corrections), passing templates or skeleton code, and, rarely, manual script or configuration updates. 
For model \textbf{performance}, we use the Dice coefficient~\cite{zijdenbos1994morphometric} for segmentation and accuracy for classification, both measured on test sets using identical evaluation protocols. For implementation \textbf{quality}, we analyse structural designs including preprocessing strategies, data augmentation techniques, neural network architectures, and computational efficiency (GPU memory usage and training time). On selected datasets, we conduct detailed comparisons to assess whether generated pipelines adhere to medical imaging practices and properly handle volumetric data properties.
All experiments are conducted with Pass@3, running each configuration three times and reporting the best-performing trial. 

\subsection{Experimental results}
\label{sec:results}
\subsubsection{Limitations of Existing Code Generation Approaches}
\label{sec:rq1_limitations}
To identify the fundamental limitations of existing code generation approaches on medical imaging tasks (RQ1), Table~\ref{fig:baseline_evaluation} presents the evaluation of existing baseline coding tools and LLMs on benchmarking medical image processing tasks.

\textbf{Proprietary IDE tools} require 5--17 human interventions across all tasks, revealing a fundamental inability to autonomously handle domain-specific requirements.
Interventions predominantly take the form of design guidance and preprocessing specification, such as selecting appropriate architectures, applying correct preprocessing, and accounting for volumetric data properties, as well as error correction for frequent execution failures. 
While these tools can generate syntactically correct code, they lack domain knowledge to make appropriate design decisions, such as selecting appropriate architectures, applying appropriate preprocessing, and accounting for volumetric data properties, leading to frequent failures that require human correction. Notably, Antigravity performs most competitively (0.6372--0.9938, fewer than 10 interventions) owing to its partial memory retention across sessions, indirectly suggesting the value of cross-project knowledge accumulation.
\textbf{Open-source LLMs} exhibit severe limitations in both domain knowledge and autonomous error correction.
Interventions span the broadest range, including domain design guidance, preprocessing specification, manual script execution, and error message relay. 
Without domain guidance, these models make critical preprocessing errors that break fundamental medical imaging properties. 
For example, GPT-oss applies unrestricted Resize instead of voxel-space Resample, corrupting physical scale information and causing near-zero segmentation performance (0.0067 Dice on FLARE22), yet lacks the validation mechanisms to detect and correct such errors autonomously.
\textbf{Proprietary LLMs} demonstrate stronger baseline capabilities but still require 5--12 interventions with inconsistent performance across datasets (0.5058--0.9752 Dice), revealing that general coding ability alone is insufficient for full autonomy in specialised domains even with state-of-the-art models.
Interventions are more mixed, spanning both domain design guidance and workflow instructions, reflecting stronger but still insufficient domain awareness.
\noindent\fbox{%
    \begin{minipage}{\dimexpr\linewidth-2\fboxsep-2\fboxrule}
        \textbf{Answer to RQ1}: 
       Existing code generation tools and LLMs consistently fail on medical imaging tasks due to lack of domain knowledge and validation, with Antigravity's partial memory retention suggesting cross-project knowledge accumulation as a promising direction.
    \end{minipage}%
}

\begin{table*}[!htbp]
  \centering
  \caption{Comparison of AutoMedImg Against Multi-agent and AutoML Frameworks on Medical Image Processing Tasks}
  \label{tab:vace_evaluation}
  \small
  \setlength{\tabcolsep}{2pt}
  \begin{tabular}{l|cc|cc|cc|cc|cc|cc}
    \hline
    & \multicolumn{2}{c|}{FLARE22} & \multicolumn{2}{c|}{ACDC} & \multicolumn{2}{c|}{BTCV} & \multicolumn{2}{c|}{FLARE21} & \multicolumn{2}{c|}{Promise12} & \multicolumn{2}{c}{OrganMNIST} \\
    Method & HI↓ & Dice↑ & HI↓ & Dice↑ & HI↓ & Dice↑ & HI↓ & Dice↑ & HI↓ & Dice↑ & HI↓ & Acc↑ \\
    \hline
    \textbf{Multi-agent frameworks} & & & & & & & & & & & &  \\
    \hline
    \texttt{CAMEL with GPT5.2~\cite{li2023camel}} & 6 & 0.6523 & 7 & 0.7234 & 9 & 0.6312 & 6 & 0.7018 & 6 & 0.8456 & 5 & 0.8923 \\
    \texttt{MetaGPT X~\cite{hong2023metagpt}} & 4 & 0.7132 & 3 & 0.7750 & 3 & 0.6845 & 4 & 0.7534 & 3 & 0.8821 & 3 & 0.9125 \\
    \texttt{MaCTG~\cite{zhao2024mactg}} & 3 & 0.5104 & 4 & 0.6273 & 2 & 0.5205 & 3 & 0.8414 & 2 & 0.8719 & 1 & 0.8820 \\
    \hline
    \textbf{AutoML frameworks} & & & & & & & & & & & &  \\
    \hline
    \texttt{AutoML-Agent~\cite{trirat2024automl}} & 5 & 0.7421 & 2 & 0.6907 & 4 & 0.6237 & 2 & 0.7428 & 1 & 0.8870 & 1 & 0.9752 \\
    \texttt{M$^3$builder~\cite{feng2025m}} & 6 & 0.7568 & 5 & 0.7413 & 1 & 0.7392 & 3 & 0.7464 & 3 & 0.8437 & 4 & 0.8727 \\
    \hline
    \textbf{Ours} & & & & & & & & & & & &  \\
    \hline
    \texttt{AutoMedImg-Qwen3-Coder} & 0 & 0.7309 & 0 & 0.7856 & 0 & 0.6423 & 0 & 0.8204 & 0 & 0.8696 & 0 & 0.9752 \\
    \texttt{AutoMedImg-Deepseek-V3.2} & 0 & 0.8333 & 0 & 0.7943 & 0 & 0.7678 & 0 & 0.8518 & 0 & 0.9293 & 0 & 0.9565 \\
    \texttt{AutoMedImg-GPT5.2} & 0 & 0.8090 & 0 & 0.7901 & 0 & 0.7669 & 0 & 0.8667 & 0 & 0.9396 & 0 & 0.9752 \\
    \texttt{AutoMedImg-Gemini3-pro} & 0 & \textbf{0.8440} & 0 & 0.8936 & 0 & 0.7664 & 0 & \textbf{0.8945} & 0 & 0.9347 & 0 & 0.8509 \\
    \texttt{AutoMedImg-Claude-4.5-Opus} & 0 & 0.8185 & 0 & \textbf{0.9041} & 0 & \textbf{0.7696} & 0 & 0.8789 & 0 & \textbf{0.9517} & 0 & \textbf{0.9938} \\
    \hline
  \end{tabular}
  \vspace{-8pt}
\end{table*}

\subsubsection{Performance of AutoMedImg Against Advanced Multi-agent and AutoML Frameworks}
\label{sec:rq1_vace}
To demonstrate how AutoMedImg addresses the limitations (RQ2), we compare AutoMedImg against multi-agent and AutoML frameworks under identical task configurations, providing both quantitative performance evaluation (Table~\ref{tab:vace_evaluation}) and qualitative pipeline design analysis on BTCV (Table~\ref{tab:btcv_detailed}).

\textbf{Multi-agent frameworks} demonstrate improved autonomy over baseline tools but still suffer from the identified fundamental limitations. 
CAMEL lacks domain-specific guidance, resulting in basic preprocessing and augmentation strategies. 
MetaGPT X still shows performance gaps on complex tasks (0.6845 Dice on BTCV, 0.7750 on ACDC), demonstrating that code generation quality alone is insufficient without domain validation. 
MaCTG achieves the lowest intervention counts (1--4) through intra-agent validation, yet without domain knowledge its validation fails on complex multi-organ segmentation (0.5104 on FLARE22, 0.5205 on BTCV) despite strong performance on simpler tasks, confirming that multi-agent architecture alone is insufficient without domain-specific validation.
\textbf{AutoML frameworks} exhibit different but equally limiting constraints. 
AutoML-Agent requires a manually provided code skeleton for each new task type, fundamentally constraining its applicability, and its retrieval mechanism fetches neural network architectures while ignoring preprocessing requirements for complex medical data formats, increasing interventions on specialised tasks (5 on FLARE22, 4 on BTCV). M$^3$Builder relies on manually provided dataset cards and fixed templates for segmentation and classification, lacking dynamic pipeline design capability. Its template-based approach causes hallucinations on task type selection and aborted executions on tasks outside template coverage, particularly semi-supervised segmentation (6 interventions on FLARE22). Its competitive BTCV performance (0.7392 Dice) stems from predefined template coverage and forced nnU-Net backbone rather than adaptive design capability.

\textbf{AutoMedImg} addresses the identified limitations across both framework categories. 
Domain-specific knowledge bases eliminate the need for manually provided skeletons or dataset cards, enabling fully automated generation from arbitrary datasets. Multi-stage validation prevents error propagation and hallucinations that cause framework abortions. Adaptive pipeline synthesis enables cross-project knowledge reuse without the rigidity of fixed templates. AutoMedImg achieves zero human interventions across all task types including semi-supervised segmentation, while maintaining competitive or superior performance compared to all frameworks.

\begin{table*}[!htbp]
  \centering
  \caption{Detailed Method Comparison on BTCV Dataset (Supervised 3D Segmentation)}
  \label{tab:btcv_detailed}
  \small
  \setlength{\tabcolsep}{3pt}
  \begin{tabular}{@{}llllrr@{}}
    \toprule
    Method & Preprocessing & Augmentation & Model & GPU & Time \\
    \midrule
    \multicolumn{6}{@{}l}{\textbf{Multi-agent frameworks}} \\
    \midrule
    CAMEL-GPT-5.2 & Res., Norm. & Rot., Flip & UNet3D & 8.9 & 146  \\
    MetaGPT X & Ori., Res., Norm. & Patch, Rot., Flip, Int. & UNet3D+RN & 6.0 & 257 \\
    MaCTG & Ori., Res., Norm. & Rot., Noise, Flip, Int., Smooth & UNet3D & 25.5 & 203 \\
    \midrule
    \multicolumn{6}{@{}l}{\textbf{AutoML frameworks}} \\
    \midrule
    AutoML-Agent & Ori., Res., Norm. & Patch & UNet3D & 14.4 & 106  \\
    M$^3$builder & Norm., Pad & Patch, Rot., Noise, Flip, Int. & nnUNet & 15.8 & 235 \\
    \midrule
    \multicolumn{6}{@{}l}{\textbf{Ours}} \\
    \midrule
    AutoMedImg-Qwen3-coder & Ori., Res., Norm. & Patch, Rot., Flip & UNet3D & 9.2 & 318 \\
    AutoMedImg-DeepSeek3.2 & Ori., Res., Norm. & Rot., Flip, Noise, Int. & UNet3D & 16.6 & 63 \\
    AutoMedImg-GPT-5.2 & Ori., Res., Norm. & Patch, Rot., Flip, Int. & UNet3D & 12.5 & 382 \\
    AutoMedImg-Gemini3-pro & Ori., Res., Norm., Pad & Patch, Rot., Flip, Int. & SwinUNet3D & 23.1 & 533 \\
    AutoMedImg-Claude4.5-Opus & Ori., Res., Norm., Pad & Patch, Rot., Noise, Flip, Int. & UNet3D & 14.9 & 477  \\
    \bottomrule
  \end{tabular}
  \par\noindent\scriptsize
  \textit{Abbreviations:} Ori.=Orientation, Res.=Resample, Norm.=Normalize, Rot.=Rotate, Int.=Intensity, RN=ResNet.\\
  GPU refers to VRAM consumption during training in GB; Time refers to training time in minutes.
\end{table*}
\noindent\fbox{%
    \begin{minipage}{\dimexpr\linewidth-2\fboxsep-2\fboxrule}
        \textbf{Answer to RQ2}: 
While multi-agent and AutoML frameworks partially address RQ1 limitations, they introduce their own constraints including manual input dependencies and fixed templates. AutoMedImg overcomes these through domain-specific knowledge, systematic multi-stage validation, and adaptive synthesis, achieving fully automated generation where existing frameworks still require human preparation.
    \end{minipage}%
}

\subsubsection{Ablation Study}
\begin{figure}
    \centering
    \includegraphics[width=0.9\linewidth]{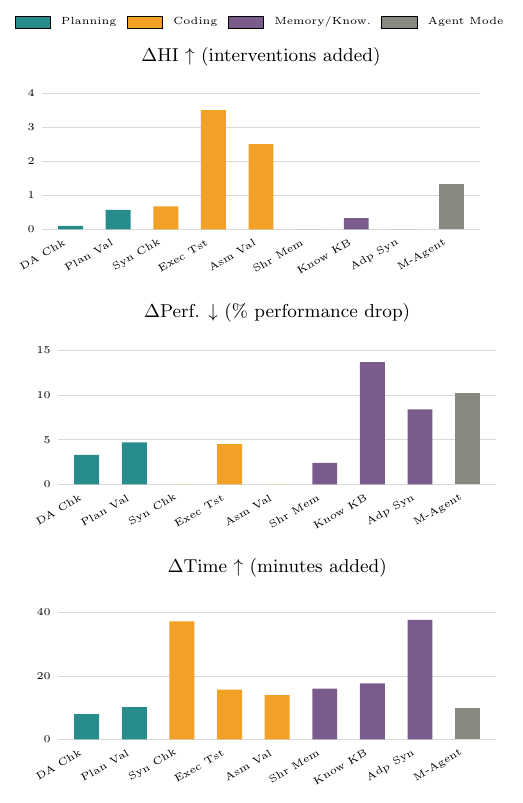}
    \vspace{-10pt}
    \caption{Ablation Study: Average Impact of Component Removal on AutoMedImg with Claude-4.5-Opus}
    \label{fig:ablation}
    \vspace{-15pt}
\end{figure}

To quantify the contribution of each individual validation and context-engineering component (RQ3), we ablate each component individually while keeping all others intact, measuring the impact on autonomy, performance, and generation efficiency across all six datasets with Claude-4.5-Opus as the backbone (Table~\ref{fig:ablation}).

\textbf{Planning Phase Validations.} Dataset analysis semantic checking removal increases interventions by 0.10 with 3.3\% performance degradation, as missing critical properties such as voxel spacing or intensity range lead to subtly incompatible downstream designs that only manifest as performance degradation rather than outright failures. 
Planning semantic validation removal has a more severe impact (+0.57 HI, -4.7\%), reflecting the central role of architecture validation in the overall pipeline — without contextual validation, agents may select fundamentally incompatible architectures such as 2D networks for 3D volumetric data.
Without formal semantic validation, occasional specification-level inconsistencies that contextual checks cannot detect go unresolved. 
While medical imaging pipelines follow straightforward sequential structures that rarely violate process ordering properties, formal verification provides an additional correctness guarantee that catches edge cases before implementation begins.
Together, these two planning validation steps form a comprehensive correctness gate, preventing a combined +0.67 HI increase and significant performance degradation before implementation begins.

\textbf{Coding Phase Validations.} Code validations exhibit a strong impact on autonomy. 
Syntax check removal adds 0.67 interventions and increases generation time by 37.2 minutes, as downstream validations eventually catch syntax errors after attempting to execute malformed code.
This result demonstrates that early-stage checks prevent costly downstream failures.
Execution testing removal dramatically increases HI by 3.50 with 4.5\% performance degradation, representing the most critical validation for automated generation. 
Without runtime verification, functional errors, such as tensor dimension mismatches or incorrect function calls, can propagate to later stages, requiring substantial human debugging.
Assembly validation removal causes 2.50 additional interventions, confirming its necessity for detecting integration issues between individually validated modules.
\begin{table*}[!htbp]
  \centering
  \caption{Comparison of AutoMedImg with Automated Medical Imaging Methods}
  \label{tab:automation_comparison}
  \footnotesize
  \setlength{\tabcolsep}{2.5pt}
  \begin{tabular}{l c c c c c c c}
    \toprule
    Method & \makecell{Complete\\Pipeline} & \makecell{Task Specific\\Design} & \makecell{Multi-task\\Support} & \makecell{Cross-project\\Reference} & \makecell{Multi-stage\\Validation} & \makecell{Production-ready\\Implementation} & \makecell{Fully\\Automated} \\
    \midrule
    NAS~\cite{wang2024mednas} & \makecell{\xmark\\\scriptsize (NN only)} & \makecell{\xmark\\\scriptsize (Depend on\\\scriptsize search space)} & \cmark & \xmark & \xmark & \xmark & \makecell{\xmark\\\scriptsize (Predefined search\\\scriptsize space needed)} \\
    nnU-Net~\cite{isensee2021nnu} & \cmark & \cmark & \makecell{\xmark\\\scriptsize (Seg. only)} & \xmark & \xmark & \cmark & \makecell{\xmark\\\scriptsize (Manual setup\\\scriptsize needed)} \\
    MedSAM~\cite{MedSAM} & \makecell{\xmark\\\scriptsize (Inference\\\scriptsize only)} & \makecell{\xmark\\\scriptsize (Fixed pre-trained\\\scriptsize model)} & \makecell{\xmark\\\scriptsize (Seg. only)} & \xmark & \xmark & \cmark & \makecell{\xmark\\\scriptsize (Interactive\\\scriptsize inference)} \\
    \midrule
    \textbf{AutoMedImg} & \textbf{\cmark} & \textbf{\cmark} & \textbf{\cmark} & \textbf{\cmark} & \textbf{\cmark} & \textbf{\cmark} & \textbf{\cmark}\\
    \bottomrule
  \end{tabular}
  \vspace{-8pt}
\end{table*}

\textbf{Memory and Knowledge Components.} Shared memory removal causes 2.4\% performance degradation with a 16-minute time increase, indicating its role in maintaining context consistency across agents. 
Removing the knowledge base results in the largest performance degradation (-13.7\%) with minimal increase in HI (+0.33), indicating that domain-specific knowledge primarily guides appropriate designs. 
Without medical imaging practices and conventions, agents generate
syntactically correct but suboptimal pipelines, such as using torchvision transforms instead of MONAI medical imaging transforms, or applying a generic resize operation that breaks voxel spacing information. 
Adaptive synthesis removal results in 8.4\% performance degradation and an over 37-minute time overhead, confirming that cross-project learning enables agents to adapt pipelines from previous projects rather than regenerate from zero. 
Without it, agents must generate solutions independently, 
resulting in slow generation and suboptimal designs.


\noindent\fbox{%
    \begin{minipage}{\dimexpr\linewidth-2\fboxsep-2\fboxrule}
        \textbf{Answer to RQ3}: 
        AutoMedImg's validation-based context engineering achieves automated generation through three integrated mechanisms: multi-stage validation for systematic  correction, domain knowledge and shared memory for quality and efficiency, and adaptive synthesis for cross-project learning, collectively eliminating manual prompting throughout the coding process.
    \end{minipage}%
}

To isolate the architectural contribution of multi-agent design (RQ4), we compare multi-agent versus single-agent execution under identical context-engineering settings, where the single-agent mode receives the same domain knowledge, validation mechanisms, and shared memory as the full system. AutoMedImg employs a multi-agent architecture to address the fundamental limitations of single-agent systems in complex code generation: domain specialisation, context management across phases, and parallel execution. The Multi-Agent (M-Agent) category in Figure~\ref{fig:ablation} demonstrates that removing this architecture significantly degrades performance, validating the necessity of this design choice.

Experiments on AutoMedImg with Claude-4.5-Opus demonstrate that single-agent mode reduces performance by 10.2\%, requires 1.33 additional interventions, and adds 10 minutes generation time, illustrating three critical limitations:
(1) Self-validation bias occurs when a single agent possesses both generation and validation knowledge, occasionally applying validation criteria during generation then merely confirming its own decisions, allowing errors to propagate unchecked.
(2) Context compaction forces the agent to eliminate earlier validated planning outputs when exceeding context limits, causing information loss.
(3) Sequential processing adds overhead as modules must be generated sequentially.
Multi-agent architecture addresses these limitations through role-based agent separation ensuring independent oversight, phase-separated context management preserving workflow history, and concurrent execution exploiting implementation parallelism.

\noindent\fbox{%
    \begin{minipage}{\dimexpr\linewidth-2\fboxsep-2\fboxrule}
        \textbf{Answer to RQ4}: 
         Multi-agent architecture is essential for automated specialised-domain code generation. Distributed coordination enables specialised reasoning that prevents knowledge confusion, preserves context history across multi-stage workflows, and exploits parallel execution to improve workflow efficiency.
    \end{minipage}%
}
\section{Discussion}
\label{sec:discussion}

\textbf{Qualitative Comparison with Code Generation Baselines.} IDE tools and vanilla LLMs possess strong code generation capability but lack domain-specific conventions and built-in validation, resulting in suboptimal pipeline designs and substantial human correction overhead. Multi-agent frameworks provide effective inter-agent cooperation but lack systematic validation and domain grounding, limiting their effectiveness on complex domain-specific tasks. AutoML frameworks offer pipeline automation but require manually provided scaffolds for each new task type, constraining their applicability to predefined task coverage. AutoMedImg addresses these limitations by instantiating base models within a multi-agent architecture augmented with domain KBs and systematic step-wise validation, eliminating the need for manually provided scaffolds through a planning-validation workflow that dynamically synthesises task-appropriate pipelines. Furthermore, AutoMedImg distinguishes itself across all categories through cross-project knowledge accumulation via adaptive pipeline synthesis, enabling validated solutions to be retrieved and adapted for future tasks, a capability absent in all prior work.

\textbf{Qualitative Comparison with Medical Imaging Automation.}
Table~\ref{tab:automation_comparison} compares AutoMedImg with automated approaches in medical imaging and related domains. Medical imaging automation methods (NAS, nnU-Net, MedSAM) incorporate domain expertise but address only partial pipeline components, requiring manual setup or interactive inference. 
AutoMedImg is the only approach that achieves complete pipeline generation, task-specific design, multi-task support, cross-project knowledge reuse, systematic multi-stage validation and full automation.

\textbf{Threats to validity.}
Despite its advantages, AutoMedImg has certain limitations:
1) Generalisation: AutoMedImg is designed specifically for medical image processing, with domain-specific knowledge bases and agent configurations tailored to this domain. The core framework (multi-agent coordination, validation-based context propagation, shared memory, adaptive synthesis) is however domain-independent and can be extended to other domains requiring systematic dataset analysis and pipeline construction, by constructing domain-specific KBs from relevant standards, defining domain-appropriate semantic validation criteria, and adapting formal verification templates. Adapting to broader general code generation tasks beyond data analysis pipelines could also inherit the framework, but would additionally require adapting agent prompts. 
2) Generation Overhead: Multi-agent coordination with iterative validation incurs higher API costs and generation time compared to direct generation. In our experiments, a complete pipeline generation costs approximately \$20--40 with an average runtime of 32 minutes, negligible compared to manual development requiring 4--8 hours at \$200--600.
\section{Conclusion}
\label{sec:conclusion}
In this work, we present AutoMedImg, a multi-agent code generation framework for medical image processing with validation-based context engineering. By employing multi-agent coordination, multi-stage validation, and comprehensive context integration encompassing domain knowledge, shared memory, validation feedback, and adaptive pipeline synthesis, AutoMedImg achieves fully automated code generation across diverse medical imaging tasks. 
Evaluation demonstrates that AutoMedImg achieves zero human intervention while maintaining competitive performance. 
Future work will extend AutoMedImg to additional tasks such as image denoising and registration, and explore improving algorithmic performance beyond code generation.

\section{Data Availability}
The artifacts that support the results discussed in this paper are available at:
\url{https://github.com/SeanCho1996/AutoMedImg}.

\bibliographystyle{ACM-Reference-Format}
\bibliography{sample-base}

\end{document}